\let\latexaddtocontents\addtocontents
\documentclass{IOS-Book-Article}
\let\addtocontents\latexaddtocontents

\usepackage[utf8]{inputenc}

\makeatletter
\let\c@author\relax
\AtBeginDocument{\let\addtocontents\@gobbletwo\setcounter{author}{0}}
\makeatother

\usepackage{mathptmx}
\usepackage{booktabs}
\usepackage{multirow}
\usepackage{amsmath}
\usepackage{caption}
\usepackage{hyperref}
\usepackage{amsfonts}
\usepackage{makecell}
\usepackage{graphicx}
\usepackage{tabularx}
\usepackage{tablefootnote}
\makeatletter
\let\c@author\relax
\makeatother
\usepackage[backend=biber, style=numeric]{biblatex}

\begin{document}

\pagestyle{headings}

\begin{frontmatter}

\title{Stitching Gaps: Fusing \\ Situated Perceptual Knowledge \\ with Vision Transformers for \\ High-Level Image Classification}

\author[A]{\fnms{Delfina Sol} \snm{Martinez Pandiani}%
\thanks{Corresponding Author: Delfina Sol Martinez Pandiani, University of Bologna, Italy and Centrum Wiskunde en Informatica, Amsterdam; E-mail:
delfinasol.martinez2@unibo.it or dsmp@cwi.nl.}},
\author[A]{\fnms{Nicolas} \snm{Lazzari}}
and
\author[A]{\fnms{Valentina} \snm{Presutti}}

\runningauthor{Martinez Pandiani et al. 2024 (Preprint)}
\address[A]{University of Bologna, Italy}

% \markboth{Preprint, February 2024}{Preprint, February 2024}

\begin{abstract}

    \noindent \textbf{Background:} The increasing demand for automatic high-level image understanding, including the detection of abstract concepts (AC) in images, presents a complex challenge both technically and ethically. This demand highlights the need for innovative and more interpretable approaches, that reconcile traditional deep vision methods with the situated, nuanced knowledge that humans use to interpret images at such high semantic levels.

    \noindent \textbf{Objective:} To bridge the gap between the deep vision and situated perceptual paradigms, this study aims to leverage situated perceptual knowledge of cultural images to enhance performance and interpretability in AC image classification. 
    
    \noindent \textbf{Methods:} We automatically extract perceptual semantic units from images, which we then model and integrate into the ARTstract Knowledge Graph (AKG). This resource captures situated perceptual semantics gleaned from over 14,000 cultural images labeled with ACs. Additionally, we enhance the AKG with high-level linguistic frames. To facilitate downstream tasks such as AC-based image classification, we compute KG embeddings. We experiment with relative representations \cite{moschella2022relative} and hybrid approaches that fuse these embeddings with visual transformer embeddings. Finally, for interpretability, we conduct posthoc qualitative analyses by examining model similarities with training instances.

    \noindent \textbf{Results:} The adoption of the relative representation method significantly bolsters KGE-based AC image classification, while our hybrid KGE-ViT methods surpassi state-of-the-art approaches outperform existing techniques. The posthoc interpretability analyses reveal the visual transformer's proficiency in capturing pixel-level visual attributes, contrasting with our method's efficacy in representing more abstract and semantic scene elements.

    \noindent \textbf{Conclusion:} Our results demonstrate the synergy and complementarity between KGE embeddings' situated perceptual knowledge and deep visual model's sensory-perceptual understanding for AC image classification. This work suggests a strong potential of neuro-symbolic methods for knowledge integration and robust image representation for use in downstream intricate visual comprehension tasks. All the materials and code are available at {\tiny \url{https://github.com/delfimpandiani/Stitching-Gaps}}

\end{abstract}


\end{frontmatter}

\pagenumbering{arabic}

\newpage
\section{Introduction}
\label{sec:intro}
In the rapidly evolving field of Computer Vision (CV), the enduring challenge is to equip machines with human-like cognitive abilities, surpassing data-driven pattern recognition to bridge the gap between bottom-up signal processing and top-down knowledge retrieval and reasoning \cite{aditya2018explicit}. This goal is rooted in the understanding that ``humans are not mere appearance-based classifiers; we acquire knowledge from experience and language" \cite{marino2017more}. While explicit knowledge has historically been recognized as a way to improve automatic image understanding, modern data-driven techniques are rooted in the deep learning (DL) paradigm and aim to acquire the majority of this knowledge from the training data itself. 

Meanwhile, CV endeavors to address increasingly complex tasks have been proposed, including discerning abstract concepts like personality traits, political affiliations, and beauty from visual cues \cite{segalin2017socialprofiling, joo2014visualpersuasion, gray2010predictingfacial, pandiani2023seeing}. However, the limitations of the deep learning paradigm become evident in these tasks of abstract concept-based (AC) image classification, where performance remains notably low \cite{martinez2023hypericons}. 
Cognitive science suggests that ACs differ from concrete concepts in that they serve as specifiers of relations between entities, relying more on \textit{semantic} and associative relations rather than categorical distinctions \cite{crutch2006different, dunabeitia2009qualitative}. Detecting ACs, therefore, often requires integrating and inferring over perceptual information \cite{bruner1990culture, firestone2016cognition}. Indeed, abstract and cognitively complex tasks benefit from an explicit understanding of perceptual semantics, such as objects and colors \cite{martinez2021automatic}, as well as symbolic representations like common-sense associations \cite{kalanat2022symbolic} and high-level linguistic frames \cite{ciroku2023automated}. 

AC image classification emphasizes the need to complement CV models with the capacity to comprehend the relationships within a scene \cite{isola2015discovering, sadeghi2011recognition}. Innovative approaches are needed to bridge the deep learning paradigm with explicit knowledge for complex image interpretation. Key to this endeavor is the integration of resources where semantics is represented explicitly, which plays a crucial role in enhancing interpretability \cite{chen2018iterative}. This integration can be realized by combining knowledge-driven methods with data-driven methods \cite{vanbekkum2021modulardesign}. Promising results have been achieved by leveraging Knowledge Graphs (KGs) to integrate background knowledge in CV models \cite{aditya2019integrating, tiddi2022knowledge}.

Based on these insights, we introduce the ``situated perceptual knowledge" paradigm to abstract concept-based (AC) image classification. This paradigm is centered on the development of a KG integrating automatically detected perceptual semantics of images, commonsense knowledge, and ACs via the SituAnnotate \cite{martinez2023situated} ontology. We inject KG embeddings (KGE), computed on the situated KG, with image representations obtained from visual transformers through varying fusion techniques, including absolute and relative representations \cite{moschella2022relative}. We also conduct qualitative analyses to understand the models' abilities to capture symbolic and embodied aspects of image content by analyzing relevant similarities with training instances.

This work is structured as follows: In Section \ref{sec:background}, we review related work. Section \ref{sec:methods} outlines our method to construct and embed the situated perceptual ARTstract Knowledge Graph (AKG), while Section \ref{sec:experiments} presents AC image classification experiments using the embeddings. Section \ref{sec:results} presents our results, and in Section \ref{sec:discussion}, we discuss and perform post-hoc interpretation of the AC image classification results, as well as propose potential future directions. We conclude in Section \ref{sec:conclusion} with a summary of our findings and contributions.

\section{Background}
\label{sec:background}
\paragraph{High-level Image Understanding}

The field of CV aims to understand images as data \cite{hoiem2008putting, arnold2020distantviewing} and interprets scene content at various levels \cite{szeliski2022computer}, seeking high-level interpretation from visual data \cite{borji2018negative, hussain2017automatic}. Recent advancements focus on automating recognition of abstract and high-level meanings in images, including situational analysis \cite{yatskar2016situationrecognition, suhail2019mixturekernelgraph, pratt2020groundedsituation, li2017situationrecognition}, event recognition \cite{yao2019attentionawarepolarity}, and visual persuasion and intent analysis \cite{joo2014visualpersuasion, jia2021intentonomydataset, huang2016inferringvisual, guo2021detectingpersuasive}, visual sentiment analysis \cite{yao2019attentionawarepolarity, toisoul2021estimationcontinuous, achlioptas2021artemisaffective}, aesthetic analysis \cite{workman2017understandingmapping, gray2010predictingfacial}, social signal processing \cite{sun2017domainbased, li2020graphbasedsocial, li2017dualglancemodel, goel2019endtoendnetwork}, and visual rhetorical analysis \cite{ye2021interpretingrhetoric, ye2018advisesymbolismb, hussain2017automaticunderstanding}.

\paragraph{Injecting Background Knowledge in CV models}
Knowledge and reasoning have been used in CV tasks for decades now. 
Methods based on First Order Logic \cite{zhu2014reasoning, london2013collective} and Description Logic \cite{dasiopoulou2009applying} have been proposed to perform AC classification and recognition. Despite their promising results, such methods suffer from the lack of flexibility in the data representation. Images are difficult to encode in a structured form. To overcome this issue, different approaches have been proposed to exploit structured knowledge within neural networks, which are far more flexible and can work directly on the raw image. 

% Markov Logic Networks (MLN) \cite{richardson2006markov}, which uses weighted First Order Logical formulas to encode an undirected grounded probabilistic graphical model, were used by \cite{zhu2014reasoning} to reason about object affordances. Probabilistic Soft Logic also uses a set of weighted First Order Logical rules, used to declare a Markov Random Field, and has been used by \cite{london2013collective} to detect collective activities such as \emph{crossing, queuing, waiting,} and \textit{dancing} in videos. Description Logics \cite{baader2003description} models relationships between entities in a particular domain, and has been used to reason and check consistency on object-level and scene-level classification systems, such as in \cite{dasiopoulou2009applying}.

% Background knowledge can be integrated into modern CV pipelines in various ways: preprocessing knowledge to augment input, incorporating knowledge as embeddings, post-processing through explicit reasoning mechanisms, and using knowledge graphs to influence neural network architectures \cite{aditya2019integrating}. 

Most works focus on injecting knowledge from large structured resources, such as Visual Genome \cite{krishna2016visualgenome} and ConceptNet \cite{havasi2007conceptnet}. Such approach enables the creation of multi-modal architectures that are able to learn image representations that are informed by the structured resource  \cite{ektefaie2023multimodalwithgraphs}. Indeed, successful results have been obtained in various CV tasks, including image classification \cite{novack2023chils}, to visual scene recognition \cite{kalanat2022symbolic}, image captioning \cite{li2023graphadapter}, image understanding \cite{guo2019deep} and scene generation \cite{buffelli2023scenegen}. 

Combining different representations (including KG) altogether is an active research field \cite{jabeen2023multimodalreview}. Approaches that integrate the knowledge representation within the model architecture \cite{novack2023chils, li2023graphadapter} or exploit it to learn better features \cite{buffelli2023scenegen} have been proposed. Those methods, however, require an extensive training data set. Recently, with the advent of large pre-trained models, techniques to merge different approaches have been proposed. This includes techniques that unify latent spaces trained on different modalities \cite{norelli2022asif, moschella2022relative} as well as techniques that integrate different representations \cite{bollegala2022metaembedding}.

\section{Method}
\label{sec:methods}
We aim to enhance AC image classification by automatically integrating situated perceptual knowledge into image representations. This involves three steps: extracting perceptual semantic units from images (Section \ref{sec:perceptual-semantics}), integrating them with contextual knowledge into the ARTstract Knowledge Graph (Section \ref{sec:akg}), and embedding the AKG for novel image representations suited for AC classification (Section \ref{sec:embedding-akg}).

\subsection{Perceptual Semantic Units Extraction}
\label{sec:perceptual-semantics}

Cognitive neuroscience research highlights that ACs in the human brain are linked to concrete items, activating sensory-motor features associated with objects, actions, and colors \cite{hoffman2018conceptscontrol}. Additionally, emotions play a significant role in AC modeling and perception \cite{kousta2011representationabstract, vigliocco2013representation}, suggesting that ACs are grounded in tangible experiences and sensory perceptions. Building on this insight, we extract perceptual semantics (PS) as cognitive-based intermediary semantic units for image representation. These include actions, age, art style, dominant colors, evoked emotions, human presence, depicted objects, and an automatically generated image caption. Table \ref{tab:detectors_listt} provides an overview of the selected extractors, including their architectural backbones, the datasets on which they are trained, and a description of their task. They have been selected through manual investigation, focusing on easily available, off-the-shelf models trained or fine-tuned for relevant semantic units and prioritizing popularity and ratings.\footnote{\tiny The models utilized for each detection are as follows: Action detection: \url{https://huggingface.co/DunnBC22/vit-base-patch16-224-in21k_Human_Activity_Recognition}; Age Tier: \url{https://huggingface.co/nateraw/vit-age-classifier}; Art Style: \url{https://huggingface.co/oschamp/vit-artworkclassifier}); Top Colors: \url{https://github.com/lokesh/color-thief}; Emotion detection: \url{https://github.com/optas/artemis/blob/master/artemis/neural_models/image_emotion_clf.py}; Human Presence: \url{https://huggingface.co/adhamelarabawy/human_presence_classifier}; Image Captioning: \url{https://huggingface.co/Salesforce/blip-image-captioning-large model}; Object Detection: \url{https://huggingface.co/Salesforce/facebook/detr-resnet-50}.} For each extractor, we manually align the output labels to the respective nodes in ConceptNet \cite{speer2017conceptnet} (except for the textual caption). By leveraging those detectors, we reflect the interpretative capabilities of CV tools across various semantic levels, facilitating automated processing for new or unseen images without requiring human-annotated ground truths.

We extract PS features from each image in the ARTstract dataset \cite{martinez2023hypericons}, a curated collection of $14,795$ cultural images associated with abstract concepts (ACs) labels. Seven labels are defined, including \textit{comfort}, \textit{danger}, \textit{death}, \textit{fitness}, \textit{freedom}, \textit{power}, and \textit{safety}. The dataset exhibits a significant class imbalance, with labels like comfort, death, and power having higher representation, which leads to low-performance when compared to other established image classification benchmarks and is mainly composed Euro- and Western-centric perspectives, potentially introducing biases.

\begin{table}[htb]
    \centering
    \label{tab:detectors_list}
    \begin{tabularx}{\textwidth}{cccX}
        \toprule
        PS Unit & Backbone & Dataset & Description \\ \midrule
        
        Action & ViT & HAR dataset \cite{anguita2013public} & Computes the probabilities for detected actions in an image such as \textit{running}, \textit{eating}, and \textit{sleeping}. \\ 
        Age Tier & ViT & Fair Face \cite{karkkainen2021fairface} & Categorizes individuals into age groups ranging from \textit{0-2} to \textit{70+}. \\
        Art Style & ViT & ArtBench-10 \cite{liao2022artbench} & Detects artistic styles such as \textit{Art Nouveau}, \textit{Baroque}, and \textit{Expressionism}. \\ 
        Top Colors & ColorThief & \textit{N/A} & Detects up to $5$ dominant colors in an image. We convert each RGB color to the CSS3 web color with the closest Euclidean distance. If a distance $\geq 50$ is detected, the color is discarded. \\
        Emotion & Artemis \cite{achlioptas2021artemisaffective} & Artemis \cite{achlioptas2021artemisaffective} & Detects the prominent emotion in an image from nine emotion categories such as \textit{amusement}, \textit{awe}, and \textit{contentment}. \\
        Human Presence & ViT & Deep Fashion v1 \cite{liu2016deepfashion} & Detects whether a human presence is in an image. \\
        Image Caption & BLIP \cite{li2022blip} & COCO \cite{lin2014microsoft} & Generate a textual description of an image. \\
        Detected Objects & DETR \cite{carion2020end} & COCO \cite{lin2014microsoft} & Detects the objects in an image. Only objects whose probability is $\geq 0.4$ are retained. \\ 
        \bottomrule
    \end{tabularx}
    \captionof{table}{Perceptual Semantic (PS) units and their associated artificial annotators, their model backbones, pretraining datasets, and other details.} 
    \label{tab:detectors_listt}
\end{table}

\begin{figure}[!h]
    \centering
    \includegraphics[width=\linewidth]{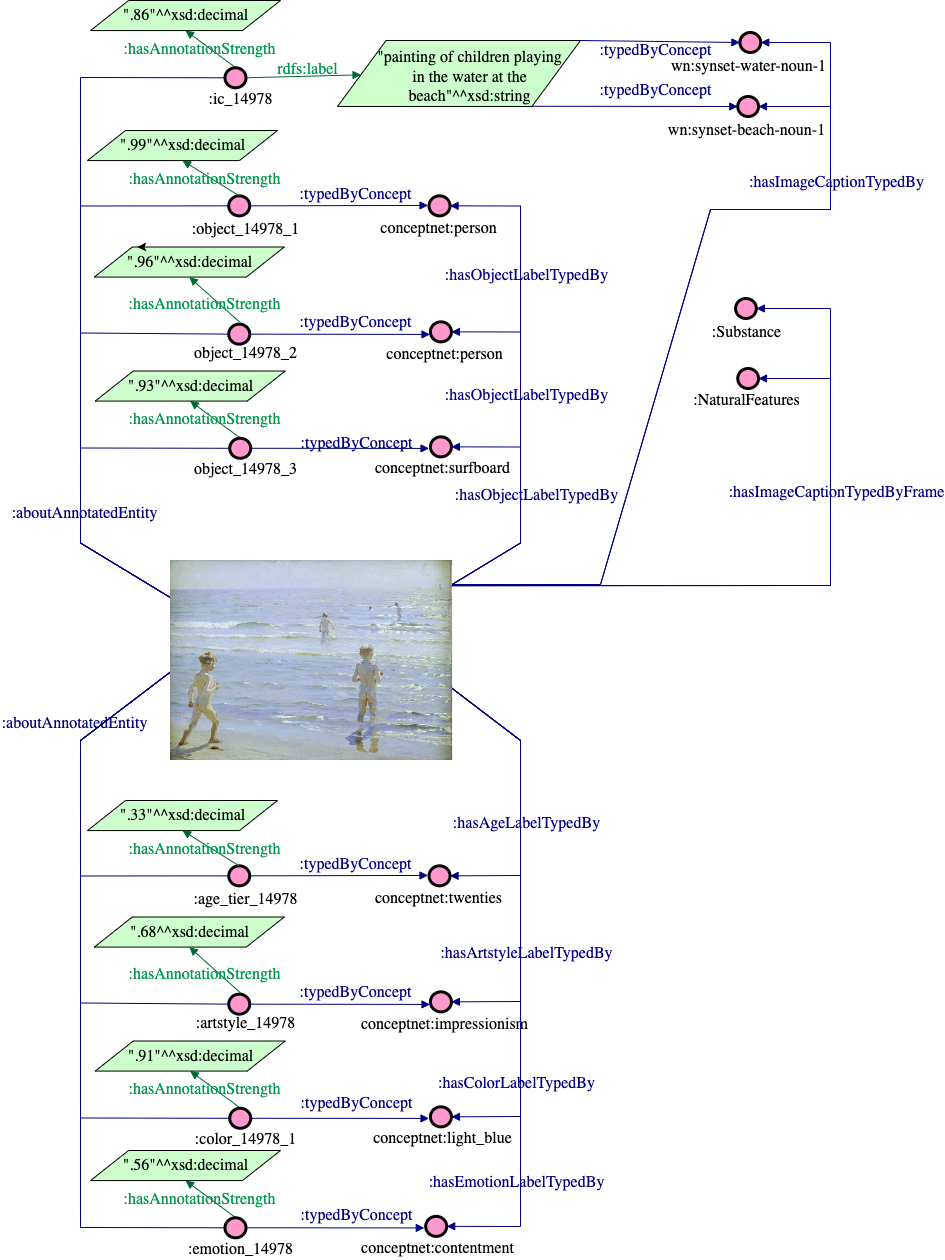}
    \caption{Subset of the A-Box of ARTstract-KG, showing the types of commonsense linguistic knowledge connected to a single image instance. Most annotations are typed by ConceptNet concepts, while the image captions are typed by WordNet concepts as well as by linguistic frames.}
    \label{img:examplebox}
\end{figure}

\newpage
\subsection{ARTstract Knowledge Graph Creation}
\label{sec:akg}

We use the \textit{SituAnnotate} ontology \cite{martinez2023situated}, which models the situated assignment of annotation labels to information objects, and includes a module tailored for \textit{image} annotation situations. To reify the PS labels, we represent each as an instance of the \textit{Annotation} class and connect it to its \texttt{AnnotationSituation}, associated \texttt{Image}, utilized \texttt{LexicalEntry}, assigned \texttt{AnnotationStrength}, label \texttt{AnnotationRole}, and the ConceptNet concept that provided its typification (see Figure \ref{img:examplebox}). To formally represent the annotation contexts, each entry row in Table \ref{tab:detectors_listt} is reified as an instance of a subclass of \texttt{ImageAnnotationSituation}. The resulting triples contain detailed information about these annotation situations, including geographical locations, timestamps, annotators, specific model architectures, datasets, and more. To further enhance the KG, following \cite{ciroku2023automated} we extract WordNet synsets from the captions using FRED \cite{gangemi2016framester}, and employ these as triggers for the extraction of high-level linguistic frames. The KG was built using RDFlib, which facilitated the mapping of PS from a JSON file to the SituAnnotate ontology, which was accessed directly via its permanent IRI\footnote{\tiny \url{https://w3id.org/situannotate}}. Additionally, we employed the Framester \cite{gangemi2016framester} schema to reference ConceptNet and WordNet IRIs.

\subsection{ARTstract Knowledge Graph Embedding}
\label{sec:embedding-akg}

In order to exploit the information encoded in the situated AKG, we compute the Knowledge Graph Embeddings of AKG by relying on TransE \cite{bordes2013translating}. KGEs transform KG components into continuous vector spaces, so as to simplify the manipulation while preserving the inherent structure of the KG \cite{wang2017knowledge}. The representation of a node associated with each image encodes all the selected PS features of an image without taking into account the raw features of the image. Since each PS is aligned to ConceptNet, the representation of two images that share the same PS feature will be similar. Before computing the embeddings, we preprocess the AKG to prevent data leakage: we remove all rows containing AC cluster names in subjects or objects.
% , removing some triples with relations: \texttt{:typedByConcept}, \texttt{:annotationWithLexicalEntry}, and \texttt{:annotationWithEvocationCluster}. 
This filtering maintains the KGEs' separation from the target AC clusters and preserves their integrity for the downstream task.

\section{Experiments}
\label{sec:experiments}
We perform AC image classification through a two-step pipeline composed of an \textit{encoding phase} that feeds a \textit{classification phase}.

\subsection{Encoding Phase}
In the encoding phase, we explore three primary approaches to represent image data: (i) rely only on the image representation produced by the KGE method (\(I_{\text{KGE}}\)); (ii) rely only on features extracted by a deep CV model (\(I_{\text{CV}}\)); (iii) combine both the KGE and CV model representations (\(I_{\text{H}}\)). When relying solely on the KGE method, we use the AKG embeddings generated with TransE. In the case of using only the CV model (\(I_{\text{CV}}\)), we evaluate three architectures: VGG \cite{simonyan2015vgg}, ResNet \cite{he2016deepresidual} and ViT \cite{zhai2022vit}. VGG and ResNet are Convolutional Neural Networks, while ViT is based on the transformer architecture. In the combined approach (\(I_{\text{H}}\)), we concatenate the KGE embedding with the most effective CV model representation, which in our case is ViT.
 
 While concatenation has demonstrated effectiveness \cite{bollegala2022metaembedding}, it merges vectors from disparate latent spaces without considering their structural differences. To address this limitation, we adopt the relative representation approach \cite{moschella2022relative}. This method constructs representations where each sample is defined in relation to a subset \(A\) of the training data \(X\), which is selected as anchor samples. Each training sample is represented with respect to the embedded anchors \(e_{a^{(j)}} = E(a^{(j)})\) with \(a^{(j)} \in A\) via a generic similarity function \(sim : \mathbb{R}^d \times \mathbb{R}^d \rightarrow \mathbb{R}\). This yields a scalar score \(r\) between two absolute representations \(r = sim(e_{x^{(i)}}, e_{x^{(j)}})\). Thus, the relative representation of \(x^{(i)} \in X\) is defined as:

\begin{equation}
r_{x^{(i)}} = (sim(e_{x^{(i)}}, e_{a^{(1)}}), sim(e_{x^{(i)}}, e_{a^{(2)}}), \ldots, sim(e_{x^{(i)}}, e_{a^{(|A|)}}))
\end{equation}

\noindent We adopt this approach to construct the relative representations of both the KGE embedding and the most effective CV model representation, ViT. Each embedding in the training distribution is represented in relation to a set of embedded anchor vectors. We randomly select 700 anchors from the training set, with 100 corresponding to each target AC class. For each image, we transform the two vector representations, \(I_{\text{KGE}}\) and \(I_{\text{ViT}}\), into their relative versions, \(I_{\text{R-KGE}}\) and \(I_{\text{R-ViT}}\) respectively. Subsequently, we apply hybrid methods to combine the relative embeddings using either concatenation ($||$) or the Hadamard product (element-wise multiplication, $\odot$). Notably, when using relative representations, the vectors maintain consistent dimensionality, eliminating the need for extension through padding.

\subsection{Classification Phase}

In the \textit{classification phase}, the output of each encoding approach is fed into a classifier. We rely on a Multi-Layer Perceptron (MLP) model to provide the final classification, comprising two sequential linear layers with a Rectified Linear Unit (ReLU) activation function and a dropout layer (dropout rate of 0.3) for regularization. Training employed the Cross-Entropy Loss with a fixed learning rate ($lr = 0.001$) and 50 epochs per architecture. The MLP model produces the probability that an image belongs to one of the $7$ labels in the ARTstract dataset. Data processing efficiency was enhanced through multi-threading with 16 workers. All the experiments have been trained in an RTX3080 with 24Gb of RAM.

\section{Results}
\label{sec:results}

% The ARTstact-KG is a publicly available resource that systematically captures and formalizes perceptual and semantic relationships within the ARTstract image dataset. Comprising over 1.9 million triples, it encompasses data from more than 14,000 unique images and offers a profound understanding of perceptual semantics in a situated way. The heart of the ARTstact-KG lies in the reification of annotation situations and perceptual semantic labels. 

\subsection{AC Image Classification Performances}

\begin{table}[!h]
    \centering
    \begin{tabular}{lccccl}
    \toprule
    {Input Embedding} & {Macro F1} & {Paradigm} \\
    \midrule
    Absolute KGE & 0.22 & SPK \\
    Absolute VGG-16 & 0.23 & DL \\
    Absolute ResNet-50 & 0.24 & DL \\
    Absolute ViT & 0.30 & DL\\
    Absolute KGE $||$ Absolute ViT & \textbf{0.31}  & Hybrid \\
    \hline
    Relative KGE & 0.27  & SPK \\
    Relative ViT & 0.28 & DL\\
    Relative KGE $\odot$ Relative ViT & 0.29 & Hybrid \\ 
    Relative KGE $||$ Relative ViT & \textit{\textbf{0.33}}  & Hybrid \\  

    \bottomrule
    \end{tabular}
    \caption{Comparison of KGE-based models with state-of-the-art DL computer vision models in terms of Macro F2 Score. The top-performing model is highlighted in both bold and italics. The second-best performing models are denoted in bold. SPK: Situated Perceptual Knowledge, DL: Deep Learning.}
    \label{tab:dl_vs_ml_vs_kg}
\end{table}

\begin{figure}[!h]
    \centering
    \includegraphics[width=.85\linewidth]{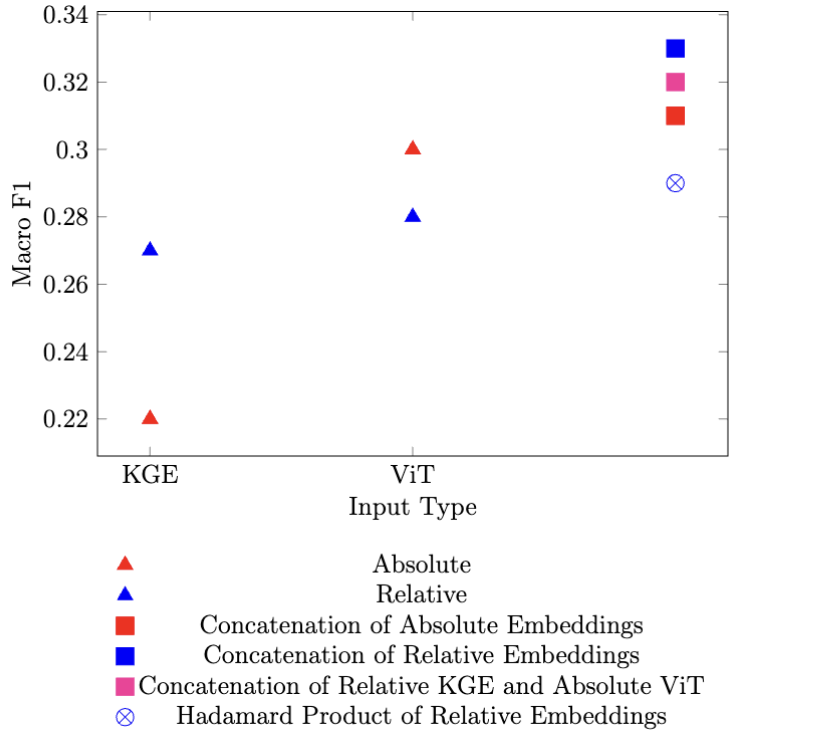}
    \caption{Macro F1 scores on the AC image classification tasks for different input embeddings.}
    \label{fig:scores}
\end{figure}

In Table \ref{tab:dl_vs_ml_vs_kg}, we present the performance metric, specifically the Macro F1 score, of our approaches compared to the state of the art. Among these models, ResNet-50 and VGG-16 achieved Macro F1 scores of 0.24 and 0.23, respectively, while ViT achieved a score of 0.30. Despite lacking access to pixel-level features, our KGE-only model demonstrated competitive results compared to the CNNs, scoring 0.22. Impressively, the Relative KGE version outperformed both CNN methods with a score of 0.27.

\paragraph{Absolute versus Relative Embeddings} RelKGE outperformed absKGE, achieving a higher Macro F1 score of 0.27 compared to 0.22 (see Figure \ref{fig:scores}). This suggests that using relative representation significantly improves KGE's performance in AC image classification. For ViT embeddings, absViT scored 0.3 in Macro F1, slightly higher than relViT at 0.28. These findings reveal a nuanced difference, indicating relative embeddings may slightly degrade ViT performance.

\paragraph{Hybrid Embeddings} Our hybrid approaches exhibited the best overall performance (as depicted in Figure \ref{fig:scores}), surpassing other methods in our study as well as the existing state of the art. Notably, the Relative KGE $||$ Relative ViT approach, which combines the two relative embeddings, achieved the highest F1 score of 0.33, representing a new benchmark for this task. Additionally, the concatenation of Absolute KGE and Absolute ViT embeddings attained a score of 0.31, further illustrating the effectiveness of hybrid methods. Lastly, the Hadamard product of the two relative embeddings scored slightly lower than Absolute ViT (0.29 versus 0.30), but it remains comparable and offers enhanced interpretability, making it a valuable option for analysis.

\section{Discussion}
\label{sec:discussion}
\subsection{The ARTstact-KG}

The ARTstact-KG is a comprehensive resource containing over 1.9 million triples derived from the ARTstract dataset, encompassing situated annotation data from more than 14,000 unique images. 
It provides detailed information about perceptual semantics, facilitated by the reification of annotation situations and semantic labels. Annotation situations capture various details such as geographical locations, timestamps, annotators, model architectures, and datasets. Similarly, semantic labels assigned to images are reified as instances of the \texttt{Annotation} class, forming connections between annotations, annotation situations, images, lexical entries, annotation strengths, annotation roles, and ConceptNet concepts, while linguistic frames extracted from image captions further enhance its expressiveness, offering a comprehensive linguistic context for each image.

\subsection{AC Image Classification}

Our results show the effectiveness of various embedding approaches in enhancing AC image classification performance. 
In our study, we implemented the relative representation method \cite{moschella2022relative}, encoding each instance relative to selected anchor points. Our results suggest that the relative representation method improves KGE-based models by providing more meaningful cluster-level semantic information, enhancing semantic resolution. Conversely, Absolute ViT outperformed Relative ViT, indicating that ViT may not benefit from the semantic bias introduced by relative representation, potentially losing fine-grained local differences and spatial resolution critical for pixel-level models like ViT. These findings underscore the relative representation method's potential to boost KGE-based image classification, offering a valuable alternative to ViT. Additionally, our hybrid embedding approaches, particularly the combination of Relative KGE and Relative ViT embeddings, showcase the highest F1 score attained in this task, setting a new benchmark for AC image classification. The competitive performance of hybrid methods underscores the effectiveness of integrating different types of embeddings to leverage their respective strengths.

\subsection{Post-Hoc Interpretability}

\subsubsection{Perceptual Disparities: ViT vs. KGE} 

\begin{figure}[!h]
    \centering
    \includegraphics[width=.9\linewidth]{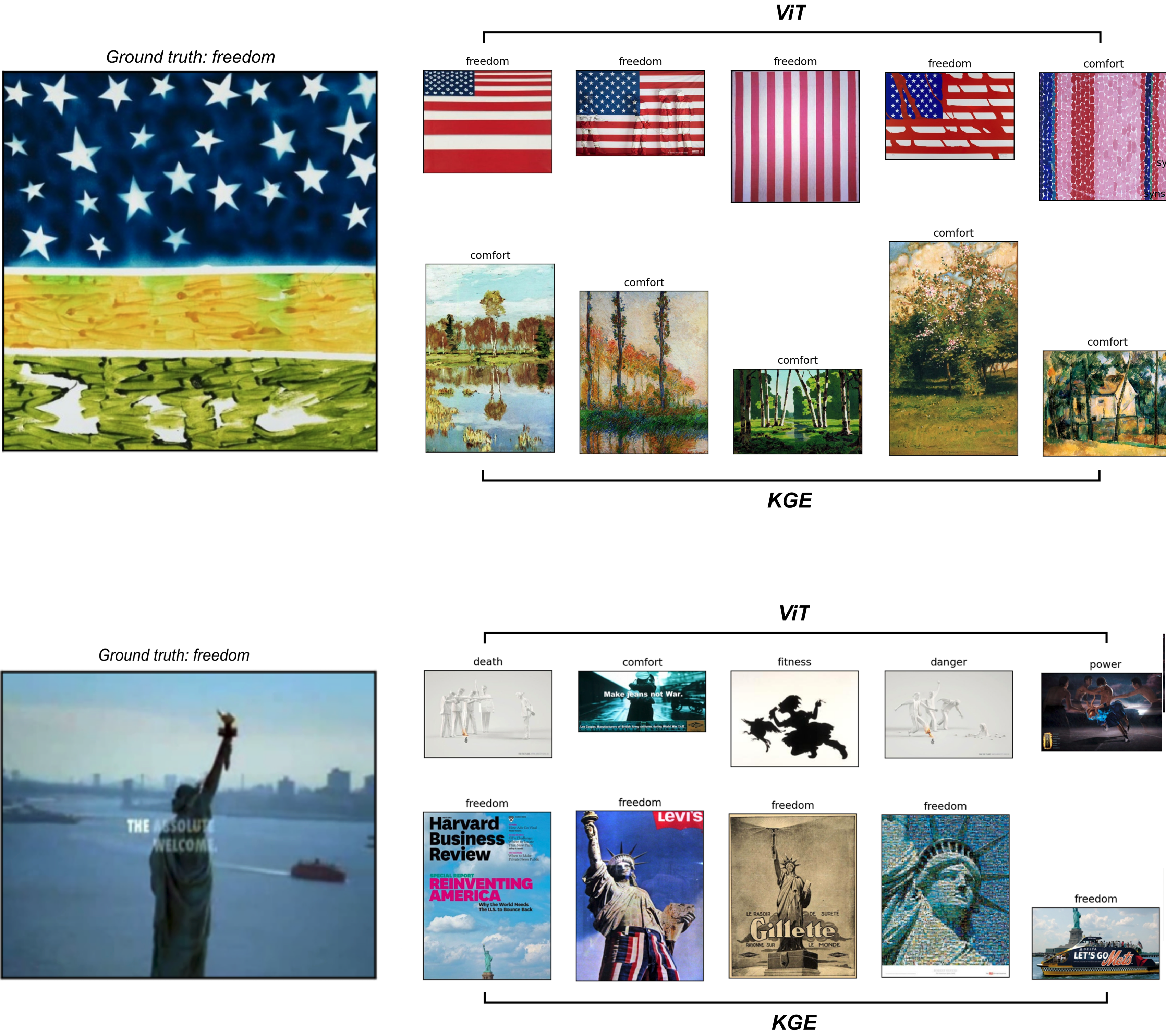}
    \caption[ViT vs. KGE embeddings capture different aspects of ARTstract images.]{Absolute ViT vs. Absolute KGE embeddings capture different aspects of ARTstract images. Top: Absolute ViT captures aspects that resemble the United States flag while KGE captures more landscape-related features, Bottom: Absolute KGE demonstrates superior semantic performance than ViT by encoding similarities with perceptually diverse visions of the Statue of Liberty}
    \label{img:abs_vs_abs}
\end{figure}

To better understand the results, we conducted a post-hoc analysis on randomly selected test images, for which we retrieved the top 5 most similar training images using both ViT and KGE embeddings, and compared them. In Figure \ref{img:abs_vs_abs}, we illustrate two examples. In the first example, when using ViT embeddings, 4 out of 5 of the top similar images correctly share the ground truth label, \textit{freedom}. These images prominently feature the United States flag, indicating that ViT's encoding accentuates features reminiscent of the flag's presence. This observation suggests potential geographical and cultural bias in ViT's training data. Contrastingly, all top similar images based on Absolute KGE embeddings share the ground truth \textit{comfort}. These images exhibit a strong visual and semantic connection with the lower portion of the test image, including elements such as grass, fields, trees, and greens. This suggests that KGE embeddings may be biased towards parts of images associated with comfort, a bias possibly inherited from the dataset itself. In the second example, none of the top images based on ViT share the correct ground truth \textit{freedom}, nor do they share evident perceptual semantics. Conversely, the top similar images based on KGE not only share the correct ground truth label but also prominently feature the Statue of Liberty. In this instance, KGE successfully associates the test image with semantically relevant training instances, whereas ViT fails to encode similarity for coherent results.

\subsubsection{High-Level Semantic Proficiency of KGE} 

\begin{figure}[!h]
    \centering
    \includegraphics[width=.9\linewidth]{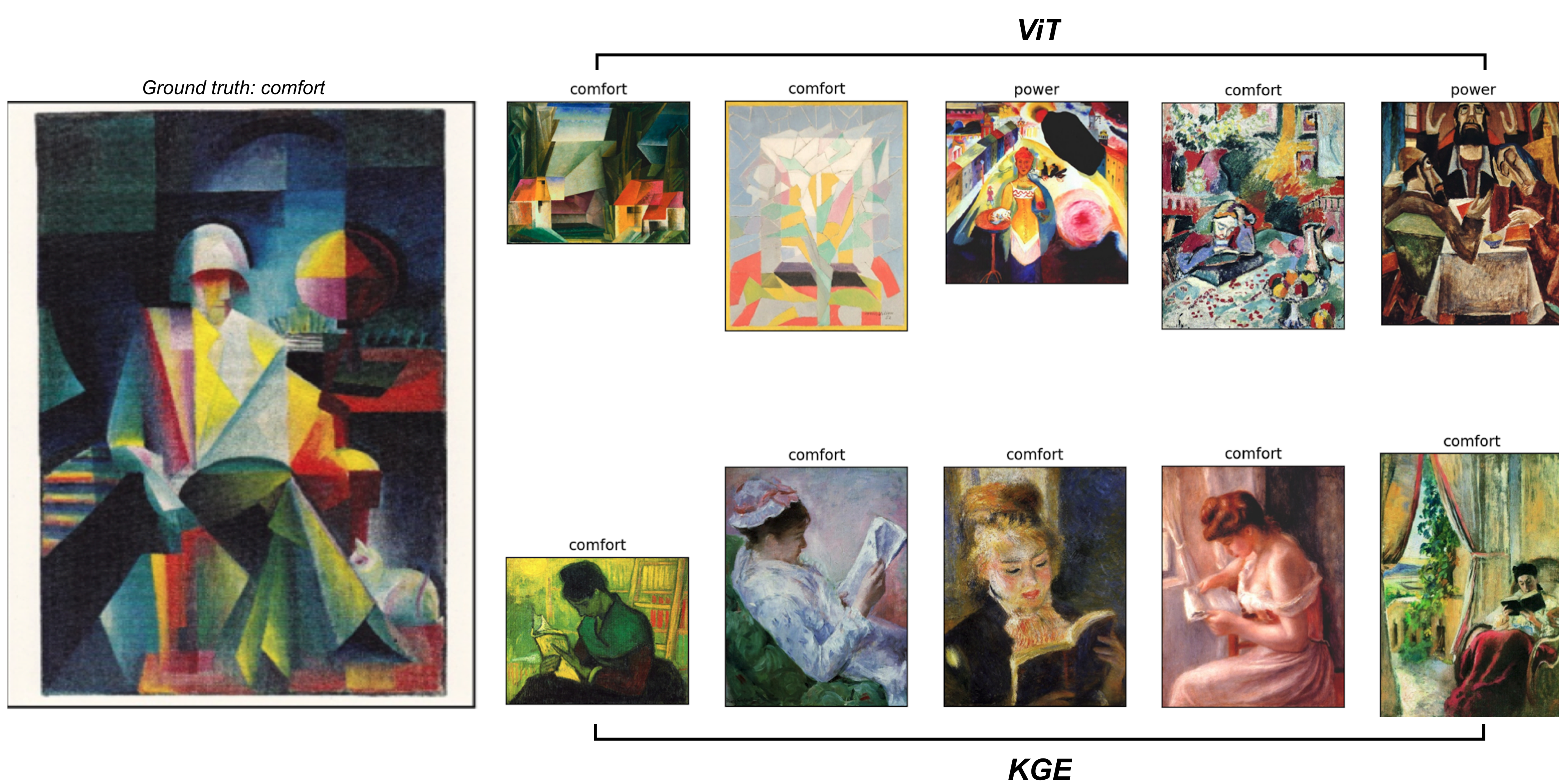}
    \caption[Contrasting semantic proficiency of Absolute KGE vs. Absolute ViT.]{Contrasting semantic proficiency of Absolute KGE vs. Absolute ViT. The top image illustrates ViT's focus on colors and textures (aesthetics), whereas KGE excels in recognizing explicit semantics, particularly women sitting on couches. In the bottom image, KGE effectively encodes the semantics of reading a book in the test artwork.}
    \label{img:reading}
\end{figure}

Further examples reveal that, even when both embeddings make correct predictions, they exhibit distinct understandings of images. Notably, KGE embeddings appear to encapsulate more ``high-level'' semantic features compared to ViT embeddings. For instance, in Figure \ref{img:reading}, the images identified as most similar by  ViT to the test image predominantly share visual characteristics reminiscent of the image's ``aesthetics" or ``style," emphasizing elements like colors, shapes, and artistic composition. However, while some of these images share the correct ground truth label \textit{comfort}, others are labeled \textit{power}. In contrast, the most similar instances identified by KGE embeddings all share the correct ground truth label \textit{comfort}, consistently conveying the same higher-level semantics--in this case, the depiction of a comfortable situation with a woman reading. KGE achieves this by aligning the top 5 similar images with depictions of women reading, whereas ViT fails to correlate the test image with any training images portraying the same scene semantics.

\begin{figure}[!h]
    \centering
    \includegraphics[width=1\linewidth]{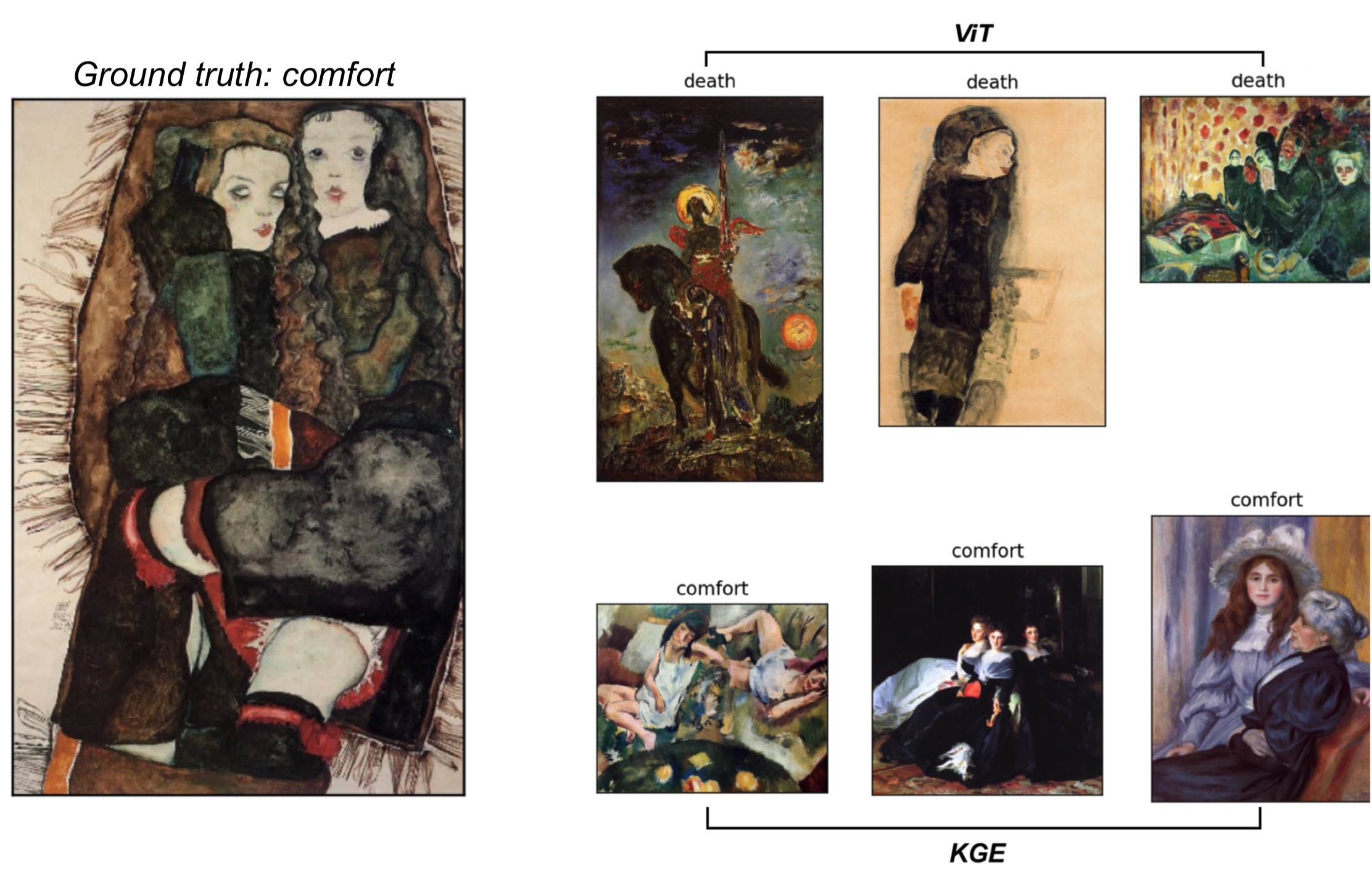}
    \caption[]{ViT misclassifies as \textit{death}, but KGE successfully associates images with crosses to the concept of \textit{comfort}, indicating ViT's focus on colors and textures.}
    \label{img:exp_couple}
\end{figure}

\begin{figure}[!h]
    \centering
    \includegraphics[width=1\linewidth]{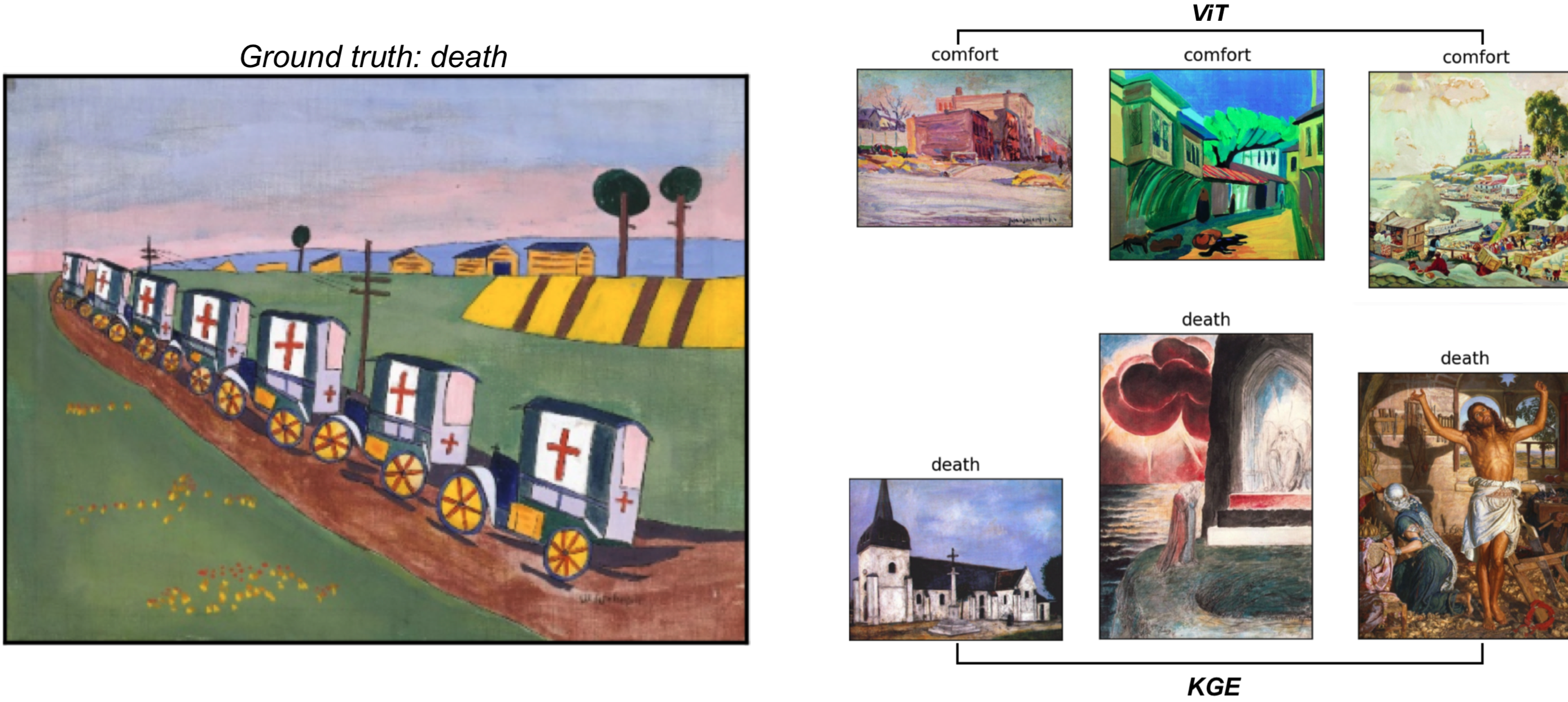}
    \caption[]{ViT misclassifies as \textit{comfort}, but KGE successfully associates images with crosses to the concept of \textit{death}.}
    \label{img:exp_death}
\end{figure}

Multiple test instances suggest that the KGE method exhibits superior performance over ViT in capturing higher-level semantics, as illustrated in Figure \ref{img:exp_couple}, a test image portraying two individuals in an intimate scenario serves as an exemplary case. While ViT-similar images primarily focus on pixel-level resemblances, such as dark colors and textures, KGE emphasizes the higher level ``situation" of individuals engaged in an intimate interaction. Notably, the majority of KGE-generated similar images depict scenes with two or more people in intimate settings, contrasting with the single individuals predominantly shown in ViT-similar images. While ViT may excel in recognizing detailed visual elements, these results suggest the KGE method's potential applicability in tasks requiring the interpretation of social interactions, relationships, or other complex high-level visual cues.

A final but compelling example of this trend is seen in Figure \ref{img:exp_death}, in which the KGE method is able to identify a ``trigger'' of a high-level semantic concept. The test image, categorized under the label \textit{death}, depicts a convoy resembling ambulances, reminiscent of those found in war zones. While ViT retrieves images primarily tagged with \textit{comfort}, likely due to the original image's warm colors, landscape composition and drawing/cartoon-like drawing features, the top similar images as based on vit feature outdoor scenes irrelevant to the ground truth of death. In contrast, the top three similar images based on KGE embeddings share the correct label of \textit{death}, evoked through the presence of crosses and crucifixion imagery. This indicates that the KGE model successfully associates images featuring crosses with the concept of death, prioritizing this connection over visual elements associated with \textit{comfort} (the ViT misclassification).

A final but compelling example highlighting this disparity is showcased in Figure \ref{img:exp_death}, where the KGE method discerns the pivotal perceptual semantic unit that acts as a``trigger" for the high-level abstract concept. The test image, categorized under the label \textit{death}, portrays a convoy resembling ambulances, reminiscent of those found in war zones. ViT retrieves images primarily tagged with \textit{comfort}, likely due to the test image's warm colors, landscape composition, and cartoon-like drawing lines; the top similar images based on ViT feature outdoor scenes irrelevant to the ground truth of \textit{death}. Conversely, the top three similar images based on KGE embeddings accurately share the \textit{death} ground truth, and all share the depiction of crosses and crucifixion imagery. This signifies the KGE model's adeptness at associating images featuring crosses with the concept of death, prioritizing this connection over visual elements associated with other ACs like \textit{comfort}.

The proficiency demonstrated by KGE embeddings is particularly remarkable considering the automated pipeline employed in constructing the ARTstract-KG. All perceptual semantic units were annotated using artificial annotators (models), indicating that they represent non-human evaluated perceptual semantics without human or manual semantic coherence checks. Consequently, this pipeline introduces inherent noise, compounded by the complexities of cultural art images, which often lack discrete objects and other detectable categories. Despite this noise, our qualitative analyses underscore the capacity of KGE embeddings to implicitly encode essential high-level semantics, a crucial aspect of our study. We believe that this discrepancy may primarily arise from the prototype selection process, wherein images are represented based on their similarity to these prototypes. Essentially, ViT's latent space heavily relies on the noise accumulated from its extensive training dataset. However, transforming this deep representation into a relative form introduces a strong prior assumption, expecting images that evoke the same AC to exhibit semantic similarity. This transformation does not perfectly align with ViT's latent space; instead, it confines the representation to specific regions within that space. This constraint potentially limits ViT's ability to express semantic relationships, as it can no longer rely solely on pixel-wise perceptual features but must effectively position images within its latent space. Consequently, the images obtained in this process may appear perplexing because the model's internal representation significantly differs from human perception. It primarily depends on subtle pixel differences, which, while effective for simple cognitive tasks, fall short in generalizing to the human internal understanding of the world.

\subsubsection{Hybridity and Complementarity} 

\begin{figure}[!h]
        \centering
        \includegraphics[width=\linewidth]{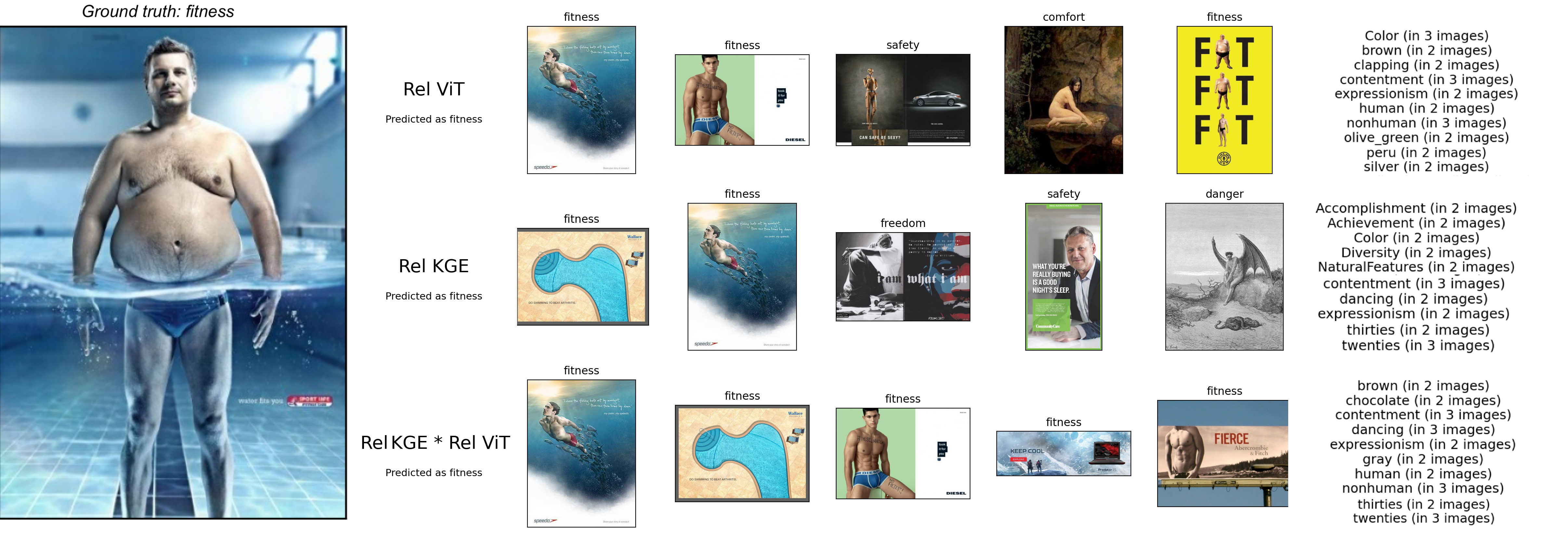}
\caption{Interpretability results for a test image labeled as \textit{fitness}. Top similar anchors are shown for the test instance using relative ViT embeddings (top row), relative KGE embeddings (middle rows), and hybrid embeddings. Shared ARTstract-KG nodes accompany each row. The hybrid embedding integrates complementary information from both relative embeddings to prioritize anchors tagged as \textit{fitness}.}
    \label{img:swimming}
\end{figure}

One critical finding was that hybrid embeddings yielded the highest classification performance, with fusing two relative embeddings resulting in more significant performance improvements than fusing two absolute embeddings. This finding highlights the complementary nature of the deep vision and situated perceptual paradigms. The results of the posthoc interpretability analyses further support this result. To illustrate, consider Figure \ref{img:swimming}, which showcases the top 5 most similar anchor images as determined by relative ViT embeddings (top row), relative KGE embeddings (middle rows), and hybrid (Hadamard product) embeddings. Each row is accompanied by the top ARTstract-KG nodes shared by each row of images, extracted via a SPARQL query on the knowledge graph. 

In this example, both relative ViT and relative KGE embeddings independently encode high similarity with anchor images sharing the correct ground truth, \textit{fitness}, and they overlap in the selection of certain anchor images. However, both embeddings also include images in their top anchor selections that do not belong to the correct target class. Remarkably, the hybrid embedding's top 5 anchors are a combination of the two unimodal correct top images, and the final selection all share the correct ground truth \textit{fitness}, combining correct anchors from each of the independent unimodal embeddings. Furthermore, we note that nodes highly shared in both single embeddings are likewise prevalent in the hybrid one. Furthermore, significant nodes exclusive to either of the embeddings are also present in the hybrid, suggesting a complementary integration of information between the embeddings.

These findings highlight the potential of utilizing the Hadamard product  ($A\circ B$) specifically on relative representations, which emphasizes anchors showing high similarity to a given image from both spatial and semantic perspectives. Through Hadamard multiplication, we assess the agreement between ViT and KGE regarding an image's similarity to prototypes, likely maintaining KGE's semantics while re-ranking images based on perceptual features detected by ViT. This operation aids in identifying anchor images with pronounced similarities to the image of interest across both embedding spaces, facilitating the recognition of anchors with dual-mode significance and unique characteristics captured by each modality. The results in Figure \ref{img:swimming} underscore the proficiency of the hybrid embedding in recognizing spatially-semantically similar anchors, attributed to the complementary nature of relative ViT and relative KGE embeddings. By combining them, we capture information sometimes missed when using them individually, offering significant benefits, particularly in situations where both pixel-level and semantic understanding are essential. The hybrid approach shows promise for various applications where understanding the underlying factors contributing to image similarity is critical.

\section{Conclusion}
\label{sec:conclusion}
This study has introduced the ARTstract Knowledge Graph (AKG) and showcased its pivotal role in advancing both interpretability and performance within the realm of AC image classification. The AKG stands out as a foundational resource, capturing perceptual semantics from over 14,000 cultural images labeled with abstract concepts (ACs), and facilitating contextual understanding in visual sense-making. By reifying perceptual semantics, encoding annotation context, and interconnecting with ConceptNet \cite{liu2004conceptnet} and Framester \cite{gangemi2016framester}, the AKG provides a robust framework for more interpretable reasoning in image analysis tasks.
Throughout this study, we have demonstrated the effectiveness of Knowledge Graph Embeddings (KGE), both absolute and relative, in enhancing AC image classification performance and post-hoc interpretability. Notably, the adoption of relative representation significantly bolstered KGE-based models, while hybrid KGE-ViT embeddings emerged as a standout performer, surpassing state-of-the-art approaches in AC image classification. Our post-hoc interpretability analyses further illuminated the strengths of different models: ViT excelled in capturing detailed pixel-level features, while KGE demonstrated proficiency in interpreting scenes and high-level semantics. The relative approach, by constraining ViT's latent space, prompted crucial considerations regarding interpretability and semantic understanding. These findings raise critical questions about the interpretability and semantic understanding of images depending on the models learning their representations. We have discussed future research avenues and the potential of hybrid approaches for cognitively complex visual tasks.

\printbibliography

\end{document}